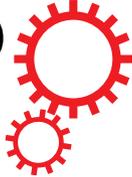

OPEN

# Forest understory trees can be segmented accurately within sufficiently dense airborne laser scanning point clouds

Hamid Hamraz[1], Marco A. Contreras[2] & Jun Zhang[1]

Airborne laser scanning (LiDAR) point clouds over large forested areas can be processed to segment individual trees and subsequently extract tree-level information. Existing segmentation procedures typically detect more than 90% of overstory trees, yet they barely detect 60% of understory trees because of the occlusion effect of higher canopy layers. Although understory trees provide limited financial value, they are an essential component of ecosystem functioning by offering habitat for numerous wildlife species and influencing stand development. Here we model the occlusion effect in terms of point density. We estimate the fractions of points representing different canopy layers (one overstory and multiple understory) and also pinpoint the required density for reasonable tree segmentation (where accuracy plateaus). We show that at a density of ~170 pt/m² understory trees can likely be segmented as accurately as overstory trees. Given the advancements of LiDAR sensor technology, point clouds will affordably reach this required density. Using modern computational approaches for big data, the denser point clouds can efficiently be processed to ultimately allow accurate remote quantification of forest resources. The methodology can also be adopted for other similar remote sensing or advanced imaging applications such as geological subsurface modelling or biomedical tissue analysis.

Global forests cover about 30% of the land surface of the earth, include 80% of plant biomass, and account for 75% of primary productivity of biosphere, providing essential and unreplaceable ecosystem services to humans and the life on our planet[1–4]. Airborne laser scanning (also known as light detection and ranging–LiDAR) technology has been extensively used in the past two decades to provide data at unprecedented spatial and temporal resolutions over large forested areas[5–10]. The LiDAR data, typically captured in the shape of 3D point clouds, can be processed to segment individual trees and subsequently obtain tree level information. This information is desired to improve the accuracy of forest assessment, monitoring, and management activities[11–16]. Due to penetration capability, the LiDAR data contains vertical vegetation structure within which understory trees can also be segmented[17, 18]. Although understory trees provide limited financial value and a minor proportion of total above ground biomass, they influence canopy succession and stand development, form a heterogeneous and dynamic habitat for numerous wildlife species, hence are an essential component of ecosystem functioning[19–22].

Several tree segmentation methods for LiDAR point clouds are by design unable to detect understory trees because they only consider top of vegetation or surface points[23–29]. More recent methods process the entire LiDAR point clouds to utilize all vertical structure information representing different vegetation layers. Some of these methods directly search the 3D space for tree segmentation and are generally computationally intensive[30–35]. Other methods reduce the computational load by analysing the vertical distribution of LiDAR points to layer the 3D space, and segmenting trees within the layers[36–41]. However, tree detection rate of understory trees (typically below 60%) is consistently lower than overstory trees (typically around or above 90%)[33, 41]. The major reason of this deficiency is the occlusion effect of higher vegetation layers that considerably decrease the penetration of LiDAR pulses toward lower layers. This fact results in much lower point density representing understory trees[18, 42–45]. Although variability in stand structure and terrain condition is the major factor affecting tree segmentation quality[46–48], a minimum point density is the basic requirement for reasonable segmentation of trees[49–51]. However, this basic requirement is typically not satisfied for understory trees in a dense forest due to occlusion[41, 52]. Very few studies have analysed the occlusion effect because of the density of the vegetation. Kükenbrink, *et al.*[45] have recently quantified the occlusion effect of higher canopy layer on lower layers and reported that at least 25% of canopy volume remain uncovered even in small-footprint LiDAR acquisition campaigns. They suggested increasing flight strip overlap, adding more observation angles and increasing point density, to uncover more of the canopy, yet they did not considered the occlusion effect on segmentation quality of individual trees.

[1]Department of Computer Science, University of Kentucky, Lexington, KY, 40506, USA. [2]Department of Forestry, University of Kentucky, Lexington, KY, 40506, USA. Correspondence and requests for materials should be addressed to H.H. (email: hhamraz@cs.uky.edu)

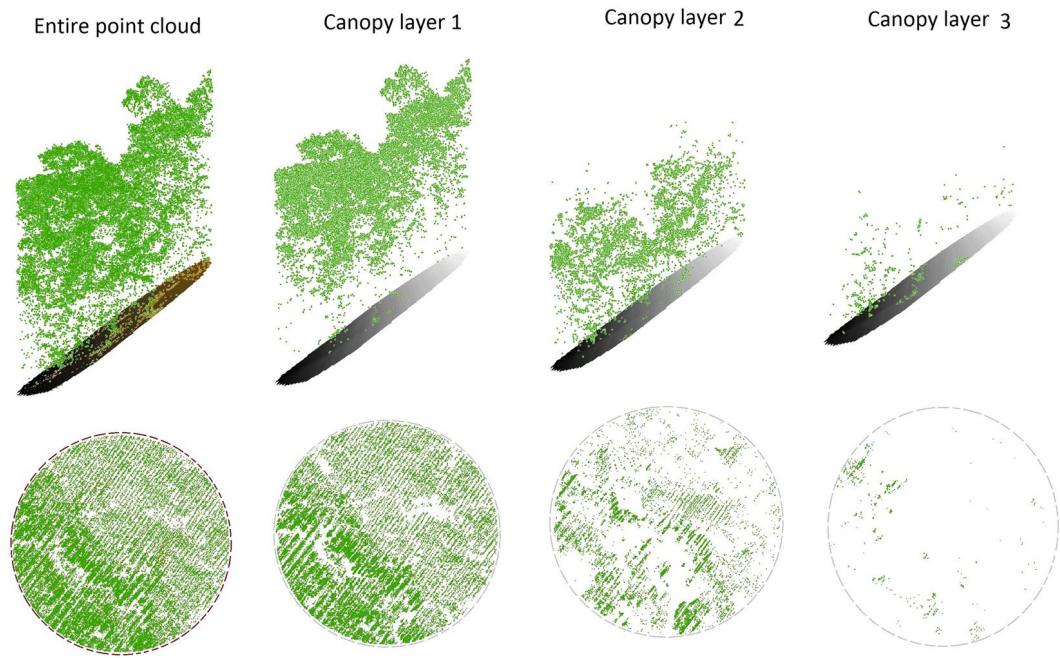

**Figure 1.** Stratification of forest LiDAR point cloud to its constituting canopy layers. Each image in the bottom row shows the aerial view of the three dimensional image right above it.

The main objective of this paper is to model the occlusion effect of higher canopy layers on lower layers in terms of point density and investigate how occlusion affects the tree segmentation quality. We define a canopy layer as a stratum of the point cloud representing tree crowns that do not overtop each other and are not over-segmented across the stratum. The top canopy layer is majorly composed of dominant and co-dominant (overstory) trees and the layers below it are majorly composed of intermediate and overtopped (understory) trees. We theorize the model by deriving a function that relates the density of the entire LiDAR point cloud (PCD) to the point density of individual canopy layers. Specifically, we present data driven analyses to: (i) estimate the fraction of LiDAR points recorded at the $n^{th}$ canopy layer that relates PCD to the point density of individual canopy layers, and (ii) estimate the required point density to reasonably segment a canopy layer by pinpointing where segmentation accuracy in the overstory canopy layer plateaus. Using (i) and (ii) in the theoretically derived function, we finally estimate the required PCD for a reasonable segmentation of understory trees–likely as accurately as overstory trees.

## Results

**Theoretical occlusion model.** Assuming all canopy layers cover the same area as the entire point cloud, PCD equals the sum of point densities of constituting canopy layers plus the density of the digital elevation model (DEM) representing the bare ground. Because the ground is different from a canopy layer in interaction with LiDAR pulses, necessitating a different density model for the DEM, we assume an infinite number of canopy layers were placed instead of the ground to simplify the analysis. Point density of the DEM approximately equals the total of point densities of the canopy layers in place of the ground. Hence PCD can be calculated as the sum of point densities of an infinite number of canopy layers (the actual ones plus those in place of the ground):

$$PCD = d_1 + d_2 + d_3 + \ldots + d_n \quad n \in \mathbb{N} \quad (1)$$

where $d_n$ denotes the point density of the $n^{th}$ canopy layer, which converges to zero as $n$ increases because point density of individual canopy layers generally decreases with proximity to ground level (Fig. 1)[41, 53, 54]. To normalize point densities, we divide both sides of Eq. 1 by *PCD*:

$$1 = p_1 + p_2 + p_3 + \ldots + p_n \quad n \in \mathbb{N} \quad (2)$$

where $p_n$ denotes the fraction of LiDAR points at the $n^{th}$ layer that can be estimated using a probability distribution function (bearing the property of summation to one).

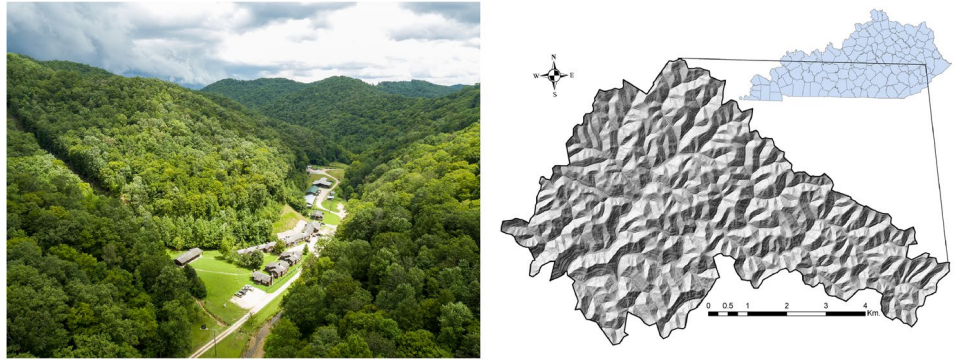

**Figure 2.** Aerial image of the camp and a glimpse over the canopy at Robinson Forest in Clayhole, KY ccaptured in August 2016 (credit: Matt Barton, Agricultural Communications Services–University of Kentucky); the forest's terrain relief map and its general location within Kentucky, USA (created using ArcMap[70] version 10.2). Robinson Forest is a ~7,400-ha natural closed-canopy deciduous forest featuring a complex dissected topography and including a diverse contiguous mixed mesophytic forest made up of approximately 80 tree species.

| Canopy Layer | Plots[1] | Starting Height (m) | | Thickness (m) | | Point Density (pt/m²) | |
|---|---|---|---|---|---|---|---|
| | | Avg. | S.D. | Avg. | S.D. | Avg. | S.D. |
| 1 | 5.86% | 15.20 | 6.56 | 8.30 | 0.81 | 44.52 | 19.02 |
| 2 | 10.17% | 3.76 | 2.80 | 8.39 | 1.20 | 7.03 | 4.29 |
| 3 | 47.50% | 0.58 | 1.08 | 6.66 | 1.38 | 0.97 | 1.01 |
| 4 | 24.71% | 0.31 | 1.12 | 6.06 | 1.54 | 0.41 | 0.83 |
| 5 | 1.76% | 0.09 | 0.08 | 5.06 | 1.35 | 0.06 | 0.54 |
| Aggregate | 90.00% | 0.31 | 0.47 | 20.93 | 9.03 | 48.09 | 23.33 |

**Table 1.** Summary statistics of canopy layers over the 50,911 sample plots regularly distributed in Robinson Forest. [1]Plots having as many number of canopy layers.

We denote the required PCD of a point cloud for a reasonable segmentation of trees forming the top canopy layer of the point cloud by $PCD_{min}$. The required PCD of a point cloud for a reasonable segmentation of trees forming the $n^{th}$ canopy layer can then be calculated using Equation 2. We hypothetically remove the $n-1$ top canopy layers of the point cloud. The resulting point cloud would have a density fraction of $1 - (p_1 + p_2 + \ldots + p_{n-1})$ of the original point cloud. Assuming this density fraction yields a density of $PCD_{min}$ for the resulting point cloud, the point density of the original point cloud for a reasonable segmentation of trees forming its $n^{th}$ top canopy layer ($pcd_{min}(n)$) by proportionality becomes:

$$pcd_{min}(n) = \frac{PCD_{min}}{1 - (p_1 + p_2 + \ldots + p_{n-1})} \qquad (3)$$

**Canopy layers and their density fractions.** We created a regularly distributed sample (40 m spacing) of 50,911 circular (radius = 15 m) plot point clouds from the entire Robinson Forest (Fig. 2) data (see Methods). We then vertically stratified each point cloud to its canopy layers (Fig. 1, see Methods). Each layer completely below a minimum height of 4 m was likely associated with ground level vegetation and was not regarded as a canopy layer. A canopy layer may however extend below this minimum height and even touch the ground. The stratification routine identified 0 layers for plots where no sufficiently large trees were present, and up to 5 layers for plots with very complex canopy structures (Table 1). Most plots had 3 (47.5%) or 4 (24.7%) canopy layers and the average number of canopy layers were 2.76. The average starting height of a canopy layer ranged from 0.1 to 15.3 m and the average thickness of a layer ranged between 5.6 and 8.4 m. Also, the average point density of a canopy layer ranged between 0.06 and 44.52 pt/m². The average starting height, thickness, and point density of the entire canopy (all layers aggregated) was 0.3 m, 20.9 m, and 48.1 pt/m², respectively. The average PCD of a plot (all canopy layers plus ground level vegetation and DEM) was 50.5 pt/m², which agrees with the point density of the initial LiDAR dataset (see Methods).

In order to estimate $p_n$ (Equation 2), we recorded a sequence of five $p_n$ values ($1 \le n \le 5$–zeros for missing layers) per each sample point cloud with at least one canopy layer. We then fitted a logarithmic series distribution[55] (having a discrete decreasing function supporting natural numbers) to all ($n$, $p_n$) pairs (N = 229,185, MSE = 0.0027 – Fig. 3):

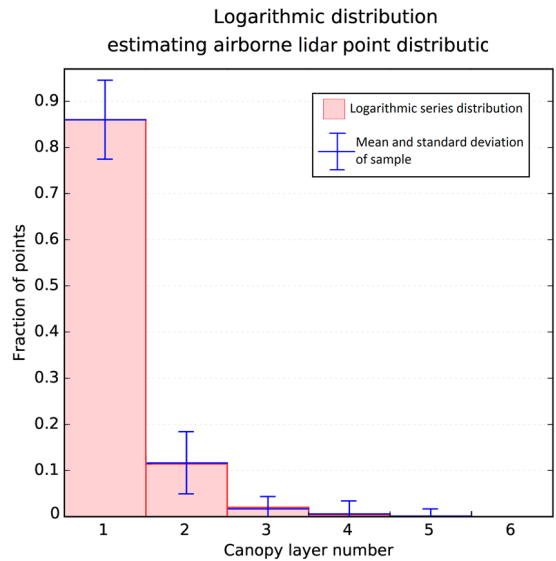

**Figure 3.** Logarithmic series distribution estimating observed fractions of LiDAR points recorded for different canopy layers. The distribution has a discrete domain supporting natural numbers.

| Plot-Level Metric | | Min | Max | Average | STD | Total | Percent of total |
|---|---|---|---|---|---|---|---|
| Slope | (%) | 10.1 | 70.5 | 42.5 | 14.8 | | |
| Aspect | | 0 | 359 | 185 | 99 | | |
| Tree count | | 6 | 27 | 13.4 | 5.1 | 303 | |
| Dominant | | 0 | 3 | 0.6 | 0.8 | 14 | 4.6 |
| Co-dominant | | 0 | 10 | 3.4 | 2.4 | 78 | 25.7 |
| Intermediate | | 2 | 10 | 5.5 | 2.5 | 126 | 41.6 |
| Overtopped | | 0 | 15 | 3.1 | 3.4 | 72 | 23.8 |
| Dead | | 0 | 5 | 0.6 | 1.0 | 13 | 4.3 |
| Species count | | 3 | 9 | 5.6 | 1.9 | 33 | |
| Shannon diversity index | | 0.80 | 2.01 | 1.47 | 0.37 | | |
| Average tree Height | (m) | 13.0 | 27.8 | 19.3 | 3.6 | | |
| Standard deviation of tree heights | (m) | 2.5 | 9.4 | 5.5 | 2.0 | | |

**Table 2.** Summary of plot level data collected from the 23 plots in Robinson Forest used for evaluating tree segmentation accuracy.

$$p_n = \frac{0.266^n}{-ln(1 - 0.266) \times n} \quad n \in \mathbb{N} \tag{4}$$

According to the derived function, for example, 86.01%, 11.44%, and 2.03% of the LiDAR points are on average returns from the first to third top canopy layers, respectively.

**Required point density for a reasonable segmentation of trees.** We decimated the point cloud to simulate a PCD of 1–50 pt/m². For each desired PCD value, we binned the point cloud into a horizontal grid with cell width of the equivalent average footprint (AFP, equals the reciprocal of square root of PCD). We then randomly selected a first return point within each cell and kept all returns associated with the LiDAR pulse generating that first return[49, 56]. For segmenting trees within the decimated point cloud, we stratified the point cloud to its canopy layers and used the surface-based method presented by Hamraz et al.[26] to segment trees within each layer. We evaluated the tree segmentation accuracy in terms of recall (measure of tree detection rate), precision (measure of correctness of the detected trees), and F-score (combined measure) (see Methods). We monitored the tree segmentation accuracy scores as a function of PCD for a sample of 23 field-surveyed plots in Robinson Forest (see Methods, Table 2) for both overstory and understory trees (Fig. 4).

As shown for overstory trees, accuracy scores are relatively stable for PCD values larger than 10 pt/m². Recall tends to decrease slightly, which is compensated by slight increases in precision resulting in a stable F-score for PCD values between 4 and 10 pt/m². Recall and consequently F-score start dropping remarkably for PCD values lower than 4 pt/m². The accuracy score trends of overstory trees concur with the previous work. As shown, the accuracy scores plateau at about 4 pt/m²[49–51], which is assumed here as the value for PCDmin. Using Equation 3, the required PCD for a reasonable segmentation of trees for as deep as three canopy layers ($pcd_{min}(3)$) would be 169.57 pt/m². This PCD approximates the required PCD to reasonably segment understory trees because they are typically found in as deep as the third canopy layer[34, 38]. Similarly, if we require a reasonable segmentation for as deep as only two canopy layers, the minimum PCD ($pcd_{min}(2)$) becomes 30.07 pt/m².

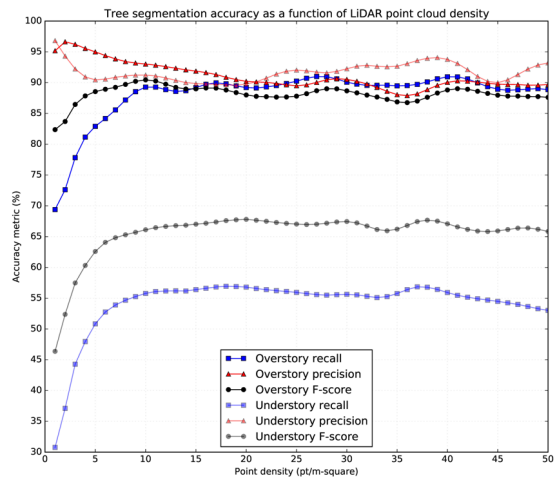

**Figure 4.** Accuracy scores of tree segmentation based on density of LiDAR point cloud for overstory and understory trees. Each symbol in the diagrams represents average across 23 sample plots from Robinson Forest.

## Discussion

As expected, the accuracy scores of understory trees are considerably lower for recall and F-score than overstory trees (Fig. 4). Directly looking at PCD is not useful for understory trees because, as mentioned, understory trees are typically found in as deep as the third canopy layer[34, 38]. Therefore, instead of PCD, we look at the effective PCD for understory trees (EUPCD), defined here as the PCD of the point cloud after removing the two top layers. The LiDAR point clouds we used had an average PCD of 50.45 pt/m², yielding 43.39 (86.01%) and 5.77 (11.44%) pt/m² for the top and the second top canopy layers using Equation 4. Removing these two layers leaves an EUPCD of 1.29 pt/m² (2.55% for the rest of layers), which is lower than the required minimum of 4 pt/m². This observation justifies the lower segmentation accuracy scores of understory trees. In fact, the understory accuracy trends in Fig. 4 only captures a domain of 0.00–1.29 pt/m² for EUPCD, which is too small to draw any conclusion for the accuracy trend of understory trees.

Considering the overstory trends (Fig. 4), a PCD of 1.29 pt/m² yielded a recall of ~70%, a precision of ~95%, and an F-score of ~82%. An EUPCD of 1.29 pt/m² should likely result in similar accuracy scores for understory trees because of similar effective PCDs. However, according to the understory trends, EUPCD of 1.29 pt/m² (equivalent to PCD of 50.45 pt/m²) consistently yielded lower scores (recall = ~55%, precision = ~93%, and F-score = ~67%), suggesting more difficulty in segmentation of inherently smaller understory trees. Quantitatively, the smaller size of understory trees resulted in ~15% lower recall, ~2% lower precision, and ~15% lower F-score than the overstory trees given similar effective low point densities. In case a similar issue rises at higher effective point densities, a potential solution would be to adjust parameters of the tree segmentation method or to use a more customized method for segmenting smaller trees at lower canopy layers, hence tightening the gap between segmentation accuracies of overstory and understory trees.

Because of the small domain of EUPCD, we based the conclusions mainly on the accuracy score trends of overstory trees. However, advancements of the LiDAR sensor technology and platforms as exemplified by the recent emergence of single-photon LiDAR[9, 57] (boosting efficiency by 10x) will enable collecting denser point clouds. Denser point clouds will not only enable effective segmentation of understory trees (as also suggested by Kükenbrink, et al.[45]), but also reveal more precise information about their segmentation accuracy trends. Denser point clouds however demand more computational resources for efficient processing. This demand has also been addressed by consistent advancements of modern computational frameworks and algorithms for big data–both for efficient storage and retrieval of big geospatial data[58, 59] as well as the parallel and distributed computing approaches for efficient processing[16, 60–62].

Lastly, different sensor and flight parameters for LiDAR acquisition can also affect the fractions of points recorded for over/understory canopy layers[51, 63]. However, point density of individual layers typically decreases with proximity to the ground[41, 53, 54]. The developed occlusion model is thus a reasonable estimator for an average case and can be consulted for future LiDAR acquisition campaigns. Moreover, performing similar analysis for different forest datasets can straightforwardly be accomplished to develop site-specific equations. As a future work, a small-footprint leaf-off dataset may be considered to create a leaf-off occlusion model in a similar manner.

## Conclusions

Airborne LiDAR data representing forested areas contain a wealth of information about horizontal and vertical vegetation structure. This information can be used to segment individual trees and subsequently retrieve morphological attributes even from understory canopy layers. Existing tree segmentation methods are unable to detect understory trees as effectively as overstory trees. This inability is mainly due to the insufficient number of LiDAR returns captured from the lower canopy levels because of the occlusion effect of the higher levels. We modelled the LiDAR occlusion effect of higher canopy layers to estimate the required PCD for reasonable segmentation of trees within lower canopy layers. We showed that a PCD of 170 pt/m² is required to reasonably segment understory trees found as deep as the third canopy layer. More accurate remote quantification of understory trees along with overstory trees will undoubtedly facilitate monitoring, management and conservation efforts.

The presented modelling methodology can also be adopted in other applications that utilize remote sensing or advanced imaging techniques, dealing with signal attenuation and/or decreased sampling. Examples of such applications include geological subsurface modelling or biomedical tissue analysis. The derived models can be used to make estimations about the potential capabilities of the associated technologies or to perform cost/utility assessment.

## Methods

### Study site.
The study site was the University of Kentucky's Robinson Forest (RF, Lat. 37.4611, Long. -83.1555) located in the rugged eastern section of the Cumberland Plateau region of southeastern Kentucky in Breathitt, Perry, and Knott counties (Fig. 2). The terrain across RF is dissected with many intermittent streams[64], slopes that are moderately steep ranging from 10 to over 100% facings predominately northwest and south east, and with elevations ranging from 252 to 503 meters above sea level. Vegetation is composed of a diverse contiguous mixed mesophytic forest made up of approximately 80 tree species with northern red oak (*Quercus rubra*), white oak (*Quercus alba*), yellow-poplar (*Liriodendron tulipifera*), American beech (*Fagus grandifolia*), eastern hemlock (*Tsuga canadensis*) and sugar maple (*Acer saccharum*) as overstory species. Understory species include eastern redbud (*Cercis canadensis*), flowering dogwood (*Cornus florida*), spicebush (*Lindera benzoin*), pawpaw (*Asimina triloba*), umbrella magnolia (*Magnolia tripetala*), and bigleaf magnolia (*Magnolia macrophylla*)[64, 65]. Average canopy cover across RF is about 93% with small opening scattered throughout. Most areas exceed 97% canopy cover and recently harvested areas have an average cover as low as 63%. After extensive logging in the 1920's, RF is considered a second growth forest ranging from 80-100 years old, and is now protected from commercial logging and mining activities[66]. RF currently covers an aggregate area of about 7,400 ha and includes about 2.5 million trees ($\pm 13.5\%$) of which over 60% are understory[16, 26].

### LiDAR acquisition campaign.
The LiDAR acquisition campaign over RF was performed in the summer of 2013 during leaf-on season (May 28–30) using a Leica ALS60 sensor, which was set at 40° field of view and 200 KHz pulse repetition rate. The sensor was flown at the average altitude of 214 m above ground at the speed of 105 knots with 50% swath overlap. Up to 4 returns were captured per pulse. Using the 95% middle portion of each swath, the resulting LiDAR dataset given the swath overlap has an average density of 50 pt/m². The provider processed the raw LiDAR dataset using the TerraScan software[67] to classify LiDAR points into ground and non-ground points. The ground points were then used to create a 1-meter resolution DEM using the natural neighbor as the fill void method and the average as the interpolation method.

### Field data.
Throughout the entire RF, 23 permanent circular plots of 0.04 ha were georeferenced with a 1.2 m precision and field surveyed in the summer of 2013. Within each plot, DBH (cm), tree height (m), species, crown class (dominant, co-dominant, intermediate, overtopped), tree status (live, dead), and stem class (single, multiple) were recorded for all trees with DBH >12.5 cm. In addition, horizontal distance and azimuth from plot centre to the face of each tree at breast height were collected to create a stem map. Site variables including slope, aspect, and slope position were also recorded for each plot. Table 2 shows a summary of the plot level data. We included a 4.7-m buffer for the LiDAR point cloud over each of the 23 field-surveyed plots for capturing complete crowns of border trees.

### Canopy stratification.
We initially calculate heights above ground of LiDAR points using the DEM and then exclude the ground points. The canopy stratification routine[41] starts with binning the LiDAR points into a horizontal grid with the cell size equal to the AFP. The height histogram (bins fixed at 25 cm) of all LiDAR points within a horizontal circular locale around every individual grid cell is then analysed. The locale should include sufficiently large number of points for building an empirical multi-modal distribution but not extend very far to preserve locality. We fixed the locale radius to $6 \times$ AFP (containing about $\pi \times 6^2$ points), which is lower bounded at 1.5 m to prohibit too small locales capturing insufficient spatial structure. Analysis of the histogram of each locale includes smoothing the height histogram using a Gaussian kernel with standard deviation of 5 m to remove variability pertaining to vertical structure of a single crown. Then the salient curves in the smoothed histogram (height ranges throughout which the second derivative of the smoothed histogram is negative) are taken as the canopy layers[36, 40]. The midpoint of the top canopy layer and the canopy layer below it is regarded as the height threshold for stratifying the top layer in that cell location (Fig. 5). Using the height thresholds determined for all grid cells, the top canopy layer is removed from the point cloud and the AFP is then updated according to the density of the remainder of the point cloud. The stratification routine iterates binning the remainder of the point cloud into a grid with the cell width of the updated AFP, analysing locales of the individual grid cells, and removing layers until the point cloud is emptied.

As the height thresholds for removing the top canopy layer in each iteration of the routine are determined using overlapping locales, the canopy layer smoothly adjusts to incorporate the vertical variability of crowns to minimize under/over-segmenting tree crowns across the layers (Fig. 1). Starting height and thickness of a canopy layer reported in Table 1 are defined as the median over all grid cells used to remove the layer from the point cloud.

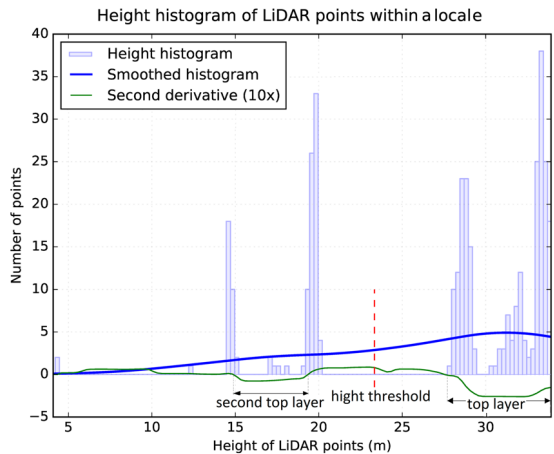

**Figure 5.** Height histogram of over a hundred LiDAR points within a circular locale used for finding the height threshold for stratifying the top canopy layer in the centre of the locale.

**Tree segmentation evaluation.** To evaluate tree segmentation, a score to each pair of LiDAR-derived tree location, assumed to be the apex of the segmented crown, and stem location measured in the field is assigned. The score is based on the tree height difference, which should be less than 30%, and the leaning angle between the crown apex and the stem location, which should also be less than 15° from nadir. The set of pairs with the maximum total score where each crown or stem location appears not more than once is selected using the Hungarian assignment algorithm and is regarded as the matched trees[26, 68]. The number of matched trees (MT) is an indication of the segmentation quality. The number of unmatched stem locations (omission errors–OE) and unmatched LiDAR-derived crowns apexes of which are not in the buffer area (commission errors–CE) indicate under- and over-segmentation, respectively. The tree segmentation accuracy metrics we used here were Recall (Re), Precision (Pr), and F-score (F), which are calculated using the following equations[69]:

$$Re = \frac{MT}{MT + OE} \tag{5}$$

$$Pr = \frac{MT}{MT + CE} \tag{6}$$

$$F = \frac{2 \times Re \times Pr}{Re + Pr} \tag{7}$$

**Data availability.** The datasets generated during and/or analysed during the current study are available from the corresponding author on reasonable request.

### Acknowledgements

This work was supported by: (1) the Department of Forestry at the University of Kentucky and the McIntire-Stennis project KY009026 Accession 1001477, (ii) the Kentucky Science and Engineering Foundation under the grant KSEF-3405-RDE-018, and (iii) the University of Kentucky Centre for Computational Sciences. The authors would also like to thank Chase Clark for helping with creation of Figure 1 and specially Prof. Dr. Kenneth L. Calvert for providing funds for open access publication.

### Author Contributions

H.H. designed the experiments, developed required computer programs, analysed the data, and composed the manuscript. M.C. provided the data, advised with the research, and contributed in composing the manuscript. J.Z. supervised the research project and helped improve the writing.

### Additional Information

**Competing Interests:** The authors declare that they have no competing interests.

**Publisher's note:** Springer Nature remains neutral with regard to jurisdictional claims in published maps and institutional affiliations.